\documentclass[letterpaper]{article} 
\usepackage{aaai24}  
\usepackage{times}  
\usepackage{helvet}  
\usepackage{courier}  
\usepackage[hyphens]{url}  
\usepackage{graphicx} 
\usepackage{natbib}  
\usepackage{caption} 
\usepackage{algorithm}
\usepackage{algorithmic}
\usepackage{subcaption}

\usepackage{booktabs}
\usepackage{amsthm}

\usepackage{booktabs} 
\usepackage{tabularx} 
\usepackage{graphicx} 

\usepackage{xcolor}

\newtheorem*{lemma*}{Lemma}
\newtheorem{lemma}{Lemma}

\usepackage{newfloat}
\usepackage{listings}
\DeclareCaptionStyle{ruled}{labelfont=normalfont,labelsep=colon,strut=off} 
\floatstyle{ruled}
\newfloat{listing}{tb}{lst}{}
\floatname{listing}{Listing}
\usepackage[T1]{fontenc}
\usepackage{amsmath}
\usepackage{amsfonts}
\usepackage{bm}
\usepackage{multirow}
\DeclareMathOperator*{\argmax}{argmax}
\DeclareMathOperator*{\argmin}{argmin}

\title{FoX: Formation-aware exploration in multi-agent reinforcement learning}
\author{
    Yonghyeon Jo, Sunwoo Lee, Junghyuk Yeom, Seungyul Han\footnote{Corresponding author}
}
\affiliations {
    Artificial Intelligence Graduate School, UNIST, Ulsan, South Korea\\
    \{yonghyeonjo, sunwoolee, yumjunghyuk, syhan\}@unist.ac.kr
}

\begin{document}

\maketitle

\begin{abstract}
Recently, deep multi-agent reinforcement learning (MARL) has gained significant popularity due to its success in various cooperative multi-agent tasks. However, exploration remains a challenging problem in MARL due to the partial observability of the agents and the exploration space that can grow exponentially as the number of agents increases. Firstly, to address the scalability issue of the exploration space, we define a formation-based equivalence relation on the exploration space and aim to reduce the search space by exploring only meaningful states in different formations. Then, we propose a novel formation-aware exploration (FoX) framework that encourages partially observable agents to visit the states in diverse formations by guiding them to be well aware of their current formation solely based on their observations. Numerical results show that the proposed FoX framework significantly outperforms the state-of-the-art MARL algorithms on Google Research Football (GRF) and sparse Starcraft II multi-agent challenge (SMAC) tasks.
\end{abstract}

\section{Introduction}
\label{sec:intro}
In recent years, multi-agent reinforcement learning (MARL) has successfully solved various real-world application challenges such as traffic control \cite{traffic1,traffic2}, games \cite{alphastar}, and robotic controls \cite{robotic}. With its increasing popularity, many multi-agent learning algorithms have been introduced \cite{indepmarl, maddpg, QMIX}. 
The MARL algorithms can be broadly classified into three categories. First, the fully decentralized methods \cite{indepmarl,fd1} train agents independently. Second, the fully centralized methods \cite{fc1} share information among agents for more efficient training. Finally, The CTDE methods \cite{coma, vdn, QMIX} provide global information during training to alleviate the nonstationarity while maintaining scalability. As a result, MARL algorithms have achieved significant success in various applications.
\begin{figure}[!t]
    \centering

    \begin{subfigure}[b]{0.595\columnwidth}
        \centering
        \includegraphics[width=\linewidth]{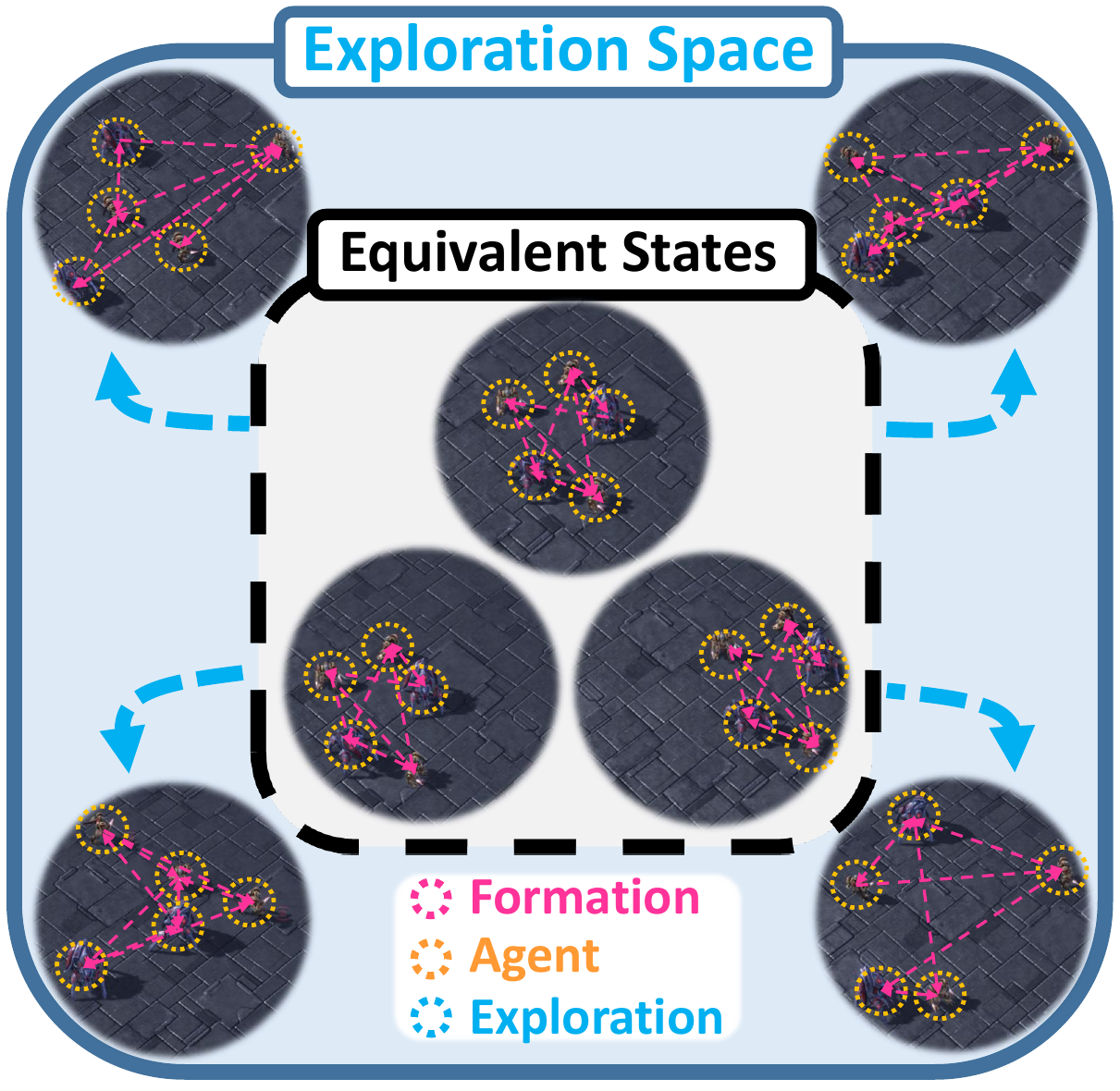}
        \caption{}
    \end{subfigure}
    \begin{subfigure}[b]{0.395\columnwidth}
        \centering
        \includegraphics[width=\linewidth]{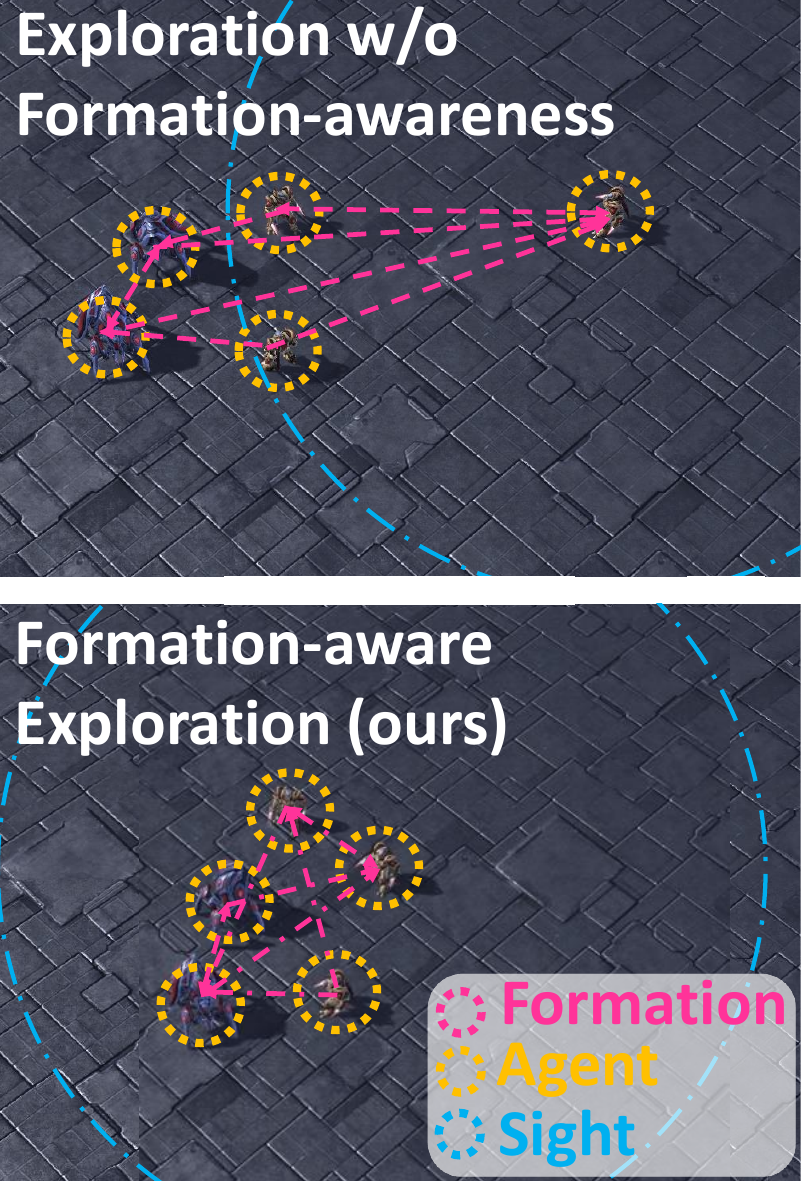}
        \caption{}
    \end{subfigure}

    \caption{(a) Illustration of formation-based state equivalence. Defining state equivalence under formations can reduce the search space efficiently. (b) Illustration of formation-awareness. Our method encourages agents to be fully aware of the formation.}
    \label{fig:concepta}
\vspace{-1em}
\end{figure}

However, MARL algorithms still face challenges from partial observability \cite{po}. Even in the traditional RL, exploration is critical to prevent the agent from converging to sub-optimal policies \cite{vc2}. To address exploration in traditional RL, exploration methods based on curiosity \cite{vc2,curi2,curi} have been proposed. Among the curiosity explorations, count-based exploration has demonstrated simple yet efficient exploration in single-agent RL problems \cite{curi,hash}. Unfortunately, the single-agent approaches may struggle in the context of MARL as partial observability makes efficient exploration even more challenging.
As consideration for agent relationships exponentially increases the search space, \cite{influence} techniques such as utilizing the influence between agents \cite{influence} or social influence \cite{social} have been introduced, but despite these efforts, efficient exploration in MARL still remains a challenge \cite{emc}.

In realistic scenarios, soccer for example, it is often impractical to consider the entire information of the field upon formulating a strategy. Rather, coaches formulate strategies based on a formation, which contains the distance information between the players in soccer games. 
For a team with high speed, a formation that allows quick transition between offense and defense would be beneficial while facing an aggressive opponent might be helpful to adopt a defensive formation.  Thus, in the soccer game, formation information is critical to winning the game against the opponents. Inspired by the real soccer game example, we define a formation in a multi-agent environment based on the differences between the agents. Based on the formation, we propose a novel formation-aware exploration (FoX) framework to visit diverse formations instead of visiting large search spaces. In FoX, we have two main contributions: 1) By defining the formation-based equivalence relation on the exploration space, we can efficiently visit states in diverse formations as illustrated in Figure \ref{fig:concepta}(a), and 2) To overcome the partial observability of agents, we design an intrinsic reward to encourage each agent to be aware of the formation viewed by the agent as shown in Figure \ref{fig:concepta}(b). In order to show the superiority of our exploration method, we conduct performance comparisons on various challenging multi-agent environments such as StarCraft II Multi-Agent Challenge (SMAC) \cite{smac}, and Google Research Football (GRF) \cite{grf}

\section{Related Work}
\subsection{Deep Multi-Agent Reinforcement Learning}
With its increasing popularity, numerous MARL algorithms \cite{indepmarl, ltc, vdn, qatten, maac, mappo} have been proposed. The fully decentralized methods \cite{indepmarl,fdec1} consider agents independently during training. To address partial observability under decentralized settings, the CTDE paradigm provides global information during training. QMIX \cite{QMIX} utilizes a mixing network for individual $Q$ values of each agent to decompose the team's expected return. QTRAN \cite{qtran} and QPLEX \cite{qplex} extend the value decomposition according to the principle of IGM. While the previous methods approach MARL problems with a value-based approach, \cite{coma, maddpg, dop, facmac} address the problem via a policy-based approach. 
\subsection{Exploration in State Space}
In environments where rewards are sparse or the state is very complex, the agent must be provided with inner motivation for efficient exploration. \cite{exp1,exp2,exp3,exp4,cbs} utilizes prediction error as intrinsic rewards for exploration. \cite{mbi,vc1,vc2,hash} designs their intrinsic rewards based on visitation count methods. While \cite{mbi} proposes using a density model as an approximator for the number of visits to a state, \cite{hash} hashes similar states to discretize the high-dimensional state space. \cite{vime,sac,maxmin,diyan}approaches exploration via information-theoretical objectives.
\subsection{Intrinsic Motivations in MARL}
As partial observability constraints efficient exploration in MARL, many studies have been conducted to address the issue \cite{liir, irat, uve, cmae}. \citep{maven} proposes maximizing the mutual information of a latent variable and agent trajectories. Using external state information such as the agents' location \cite{lazy} can greatly assist training, but assuming such information may affect generality in the application. \citep{social, influence} addresses the challenge by exploring based on the influence between the agents, \cite{roma, rode} assigns roles to the agents for task-appropriate behavior. \cite{maser} targets observations with the highest sum of individual and global value as sub-goals. On the other hand, \cite{emc} defines curiosity based on prediction errors on individual $Q$-values, whereas \cite{cds} encourages agent diversity via local trajectory.

\section{Preliminaries}
\subsection{Decentralized POMDP}
A fully cooperative multi-agent task can be seen as a decentralized partially observable Markov decision process (Dec-POMDP) \cite{dec}, represented as a tuple $G = \langle \mathcal{S}, \mathcal{A}, \mathcal{P}, r, Z, O, \mathcal{O}, I, n, \gamma \rangle$. $\mathcal{S}$ is the true state space of the environment or the observation product space of all $n$ agents, where $I$ is a finite set of $n$ agents. $\mathcal{A}$ is the set of actions. At each time step $t$, each agent $i \in I$ receives $d$-dimensional observation vector $o_t^i \in \mathcal{O}^i$ from the environment according to the observation function $O^i(s)$. Then, agents select joint action $\textbf{a}_t=(a_t^0,\cdots,a_t^{n-1})$, and the next state $s_{t+1}$ and the global reward $r_t = R(s_t,\mathbf{a}_t)$ are generated from the environment based on the transition function $P(\cdot|s_t,\mathbf{a}_t)$ and the reward function $R$. $Z$ is the latent space, and $\gamma\in [0,1)$ is the discount factor. Each agent conditions its policy $\pi^i(a_t^i|\tau_t^i)$, where $\tau_t^i = (o_0,a_0,\cdots,o_t)$ is a trajectory of $i$-th agent, as the agents choose their action based on their local observations. Individual policies $\pi_i$ form the joint policy $\mathbf{\pi}=\prod_{i=0}^{n-1}\pi^i$, and the main objective of RL is to maximize the cumulative reward sum $\mathbb{E}_{s_0,\mathbf{a_0},\cdots}[\sum_{t=0}^{T-1} r_t]$ given from the environment, where $T$ is the episode length.

\subsection{Centralized Training Decentralized Execution}
The CTDE methods provide information of the full information of agents only during training to mitigate the challenges from partial observability while maintaining decentralized execution. Among the CTDE methods are the value decomposition methods where the global value function $Q^{tot}(\bm{\tau},\textbf{a})$ is decomposed into individual values. A popular value decomposition method is QMIX \cite{QMIX}, which decomposes the global value function through a mixing network to individual agent utility functions $Q^i$ as follows:
\begin{equation}
    \argmax_{\textbf{a}} Q^{tot}(\bm{\tau},\textbf{a}) = \prod_{i=0}^{n-1}\argmax_{a^i}Q^i(\tau^i,a^i),
    \label{eq:qmix}
\end{equation}
where $\bm{\tau}=(\tau_0,\cdots,\tau_{n-1})$ is the joint trajectory. Furthermore, the individual-global-max (IGM) condition must be satisfied to maintain consistency between local and global greedy actions \cite{qtran}.
The consistency between the global and local greedy actions allows greedy local actions to result in optimal global actions.

\begin{figure}[!t]
    \centering

    \begin{subfigure}[b]{0.32\columnwidth}
        \centering
        \includegraphics[width=\linewidth, height=2cm]{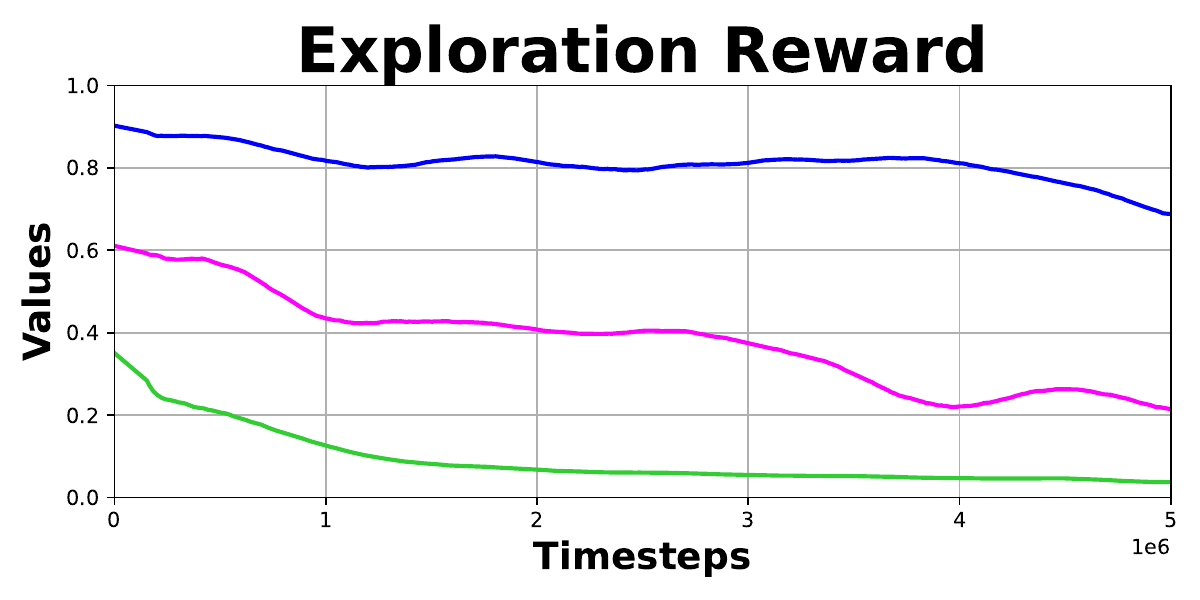}
        \caption{}
    \end{subfigure}
    \begin{subfigure}[b]{0.32\columnwidth}
        \centering
        \includegraphics[width=\linewidth]{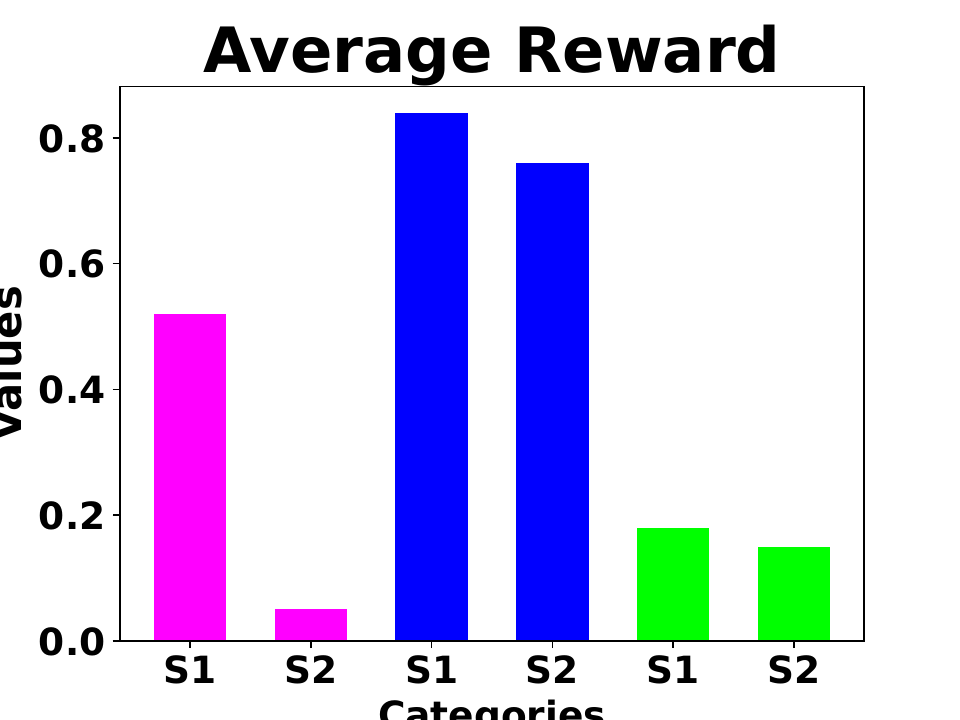}
        \caption{}
    \end{subfigure}
    \begin{subfigure}[b]{0.32\columnwidth}
        \centering
        \includegraphics[width=\linewidth]{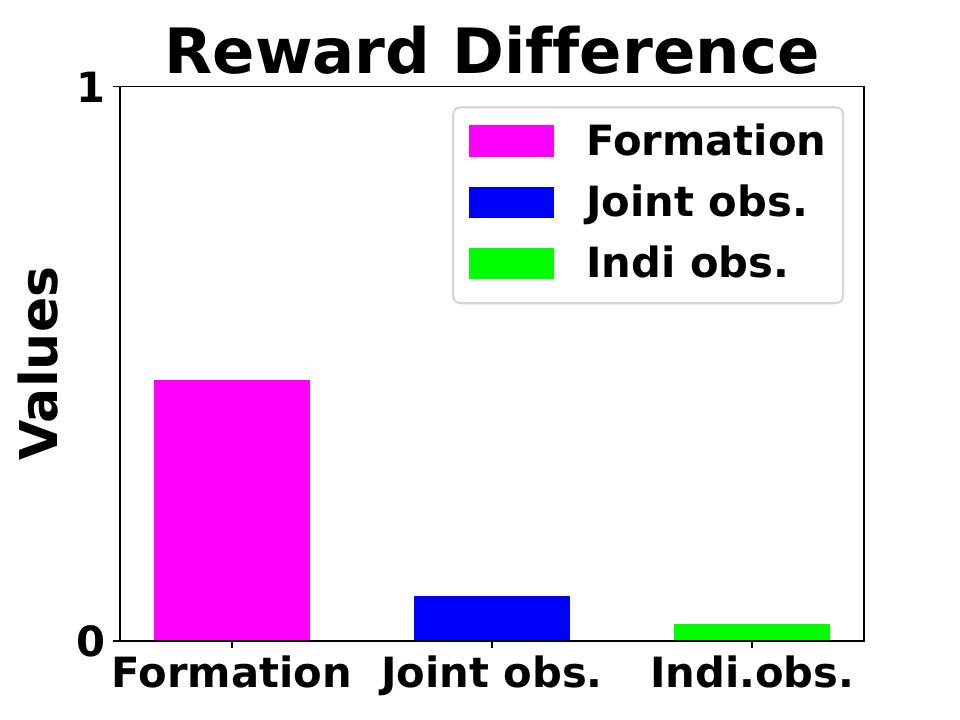}
        \caption{}
    \end{subfigure}
    \begin{subfigure}[b]{\columnwidth}
        \centering
        \includegraphics[width=\linewidth]{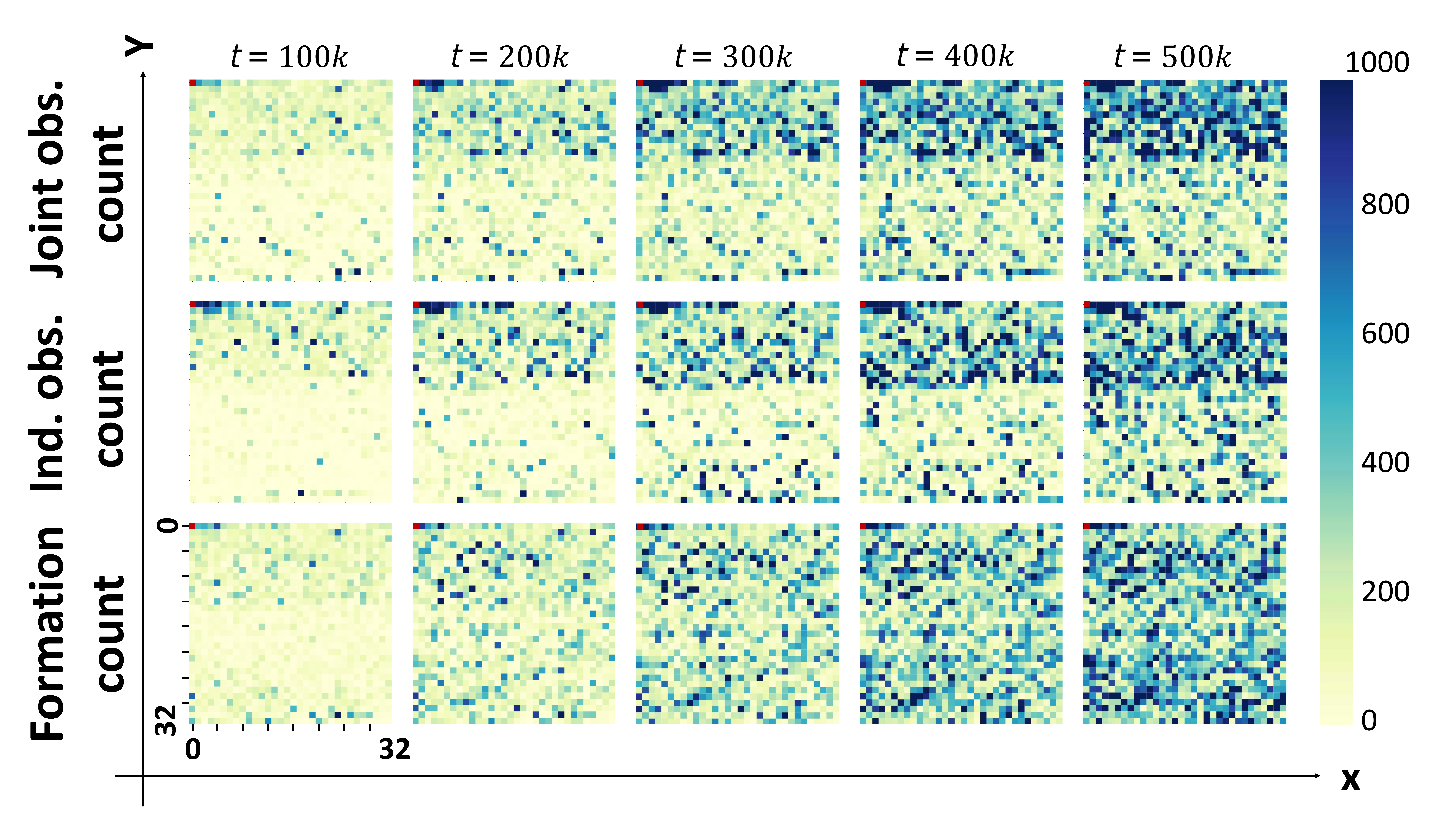}
        \caption{}
    \end{subfigure}
    \caption{(a) Pure exploration reward based on visitation count graph. (b) Average reward graph on set of rarely visited formation $S_1$ and frequently visited formation $S_2$. (c) The reward difference graph for each component. (d) Heatmap of diverse formation-based exploration space. With initial spawn formation at (0,0), a farther point in the heatmap indicates a larger difference in formations.}
    \vspace{-.5em}
    \label{fig:heatmap}
\end{figure}

\section{FoX: Formation-aware Exploration}
In this section, we introduce FoX, a novel exploration framework for cooperative multi-agent reinforcement learning. 

\subsection{Motivation of Formation-aware Exploration}
\label{sec:motiv}

In general, the assumption that agents have information about the true global state may compromise generality. Thus, in this paper, we consider a more general scenario that agents explore the observation product space, so we define an exploration space $\mathcal{S}^e$ as the observation product space $\mathcal{S}^e:=\prod_{i\in I} \mathcal{O}^i$. Then, agents should explore diverse exploration states defined as joint observations $s_e:=(o_0,\cdots,o_{n-1}) \in \mathcal{S}^e$ to experience various interactions between agents well. 

However, visiting all exploration states is a challenging problem since $\mathcal{S}^e$ grows exponentially if the number of agents and the dimension of observation space increase, and it suffers from the curse of dimensionality \cite{cod}. Thus, as motivated in real soccer games, we define the formation $\mathcal{F}$ based on the observation differences between agents, and aim to explore diverse formations instead of the entire exploration space to reduce the search space. 

In order to show the efficiency of our formation-based exploration, we conducted pure-exploration experiments in a modified 2s3z environment of SMAC \cite{smac}. Following the previous works on visitation count methods \cite{vc1}, in this setup, the agent only gets the pure exploration reward for count-based exploration $r_t \propto \frac{1}{\sqrt{N(s)}}$ to explore rarely visited components $s$, where $N(s)$ is the count for visitation on state $s$. Here, we consider three types of components $s$ for comparison: 1) joint observations  $(o^0,\cdots,o^{n-1})$, 2) individual observations $o^i$, and 3) proposed formations $\mathcal{F}$, and details of visitation count are provided in Appendix A. For all three cases, we simultaneously count the visitation of formation $N(\mathcal{F})$ during training, and compute the average $r_t$ of two state sets: a set of states that rarely visit the same formation $S_1=\{s_t\in\mathcal{S}|1\leq N(\mathcal{F}(s_t))\leq100\}$ and another set of states that frequently visit the same formation $S_2=\{s_t\in\mathcal{S}|N(\mathcal{F}(s_t))>100\}$ at $t=500k$. For efficient exploration, it is important to rapidly decrease rewards for frequently visited states $S_2$, while maintaining high rewards for novel states $S_1$.

Figure \ref{fig:heatmap} shows the result of pure exploration, where Figure \ref{fig:heatmap}(a) shows the average pure exploration reward, Figure \ref{fig:heatmap}(b) shows the average reward sets $S_1$ and $S_2$, Figure \ref{fig:heatmap}(c) shows the reward difference between the two sets $S_1$ and $S_2$, and Figure \ref{fig:heatmap}(d) illustrates the heatmap of the visitation frequency of diverse formations in the exploration space. In the heatmap, a farther point in the heatmap indicates the larger difference in formations with initial spawn formation at (0,0). From Figure \ref{fig:heatmap}(a), we can observe no significant change in reward in the case of joint observations, and rapid decrease in the reward in the case of individual observations. According to Figure \ref{fig:heatmap}(b) and Figure \ref{fig:heatmap}(c), exploration based on joint observations or individual observations is unable to highlight the difference in rewards between the set of novel states $S_1$, and the set frequently visited states $S_2$. On the other hand, in the case of our proposed formation, the reward difference between the two sets is prominently displayed, as shown in Figure \ref{fig:heatmap}(c). Such a tendency in the intrinsic rewards demonstrates that the reward decrease of our proposed formation shown in Figure \ref{fig:heatmap}(a) is indeed appropriate, and that formation-based exploration enables efficient exploration in MARL.

The efficiency of formation-based exploration is also illustrated in Figure \ref{fig:heatmap}(d). According to our heatmap, the joint observation or the individual observation-based exploration only explores a limited area of the heatmap while in the case of the individual observation, it visits a farther area in the heatmap than the joint observation case, but it still cannot visit the diverse area in the heatmap. On the other hand, we can observe that our formation-based exploration allows the agents to successfully visit more diverse areas in the heatmap. Thus, the pure exploration example shows that our method yields better exploration to visit diverse formations in exploration space.

\subsection{Formation Arrangement}

In order to reduce the search space, we aim to define a formation based on the observation differences $o^i-o^j$ between $i$-th agent and $j$-th agent, as explained in previous sections. However, The dimension of the observation difference $o^i-o^j$ is still so large that it makes exploration difficult, we first reduce the dimension of the observation difference $o_i-o_j$ as a 2-dimensional vector $D^{ij} := (||o^i-o^j||_2,~c(o^i-o^j)) \in \mathbb{R}\times \mathbb{Z}$, where $||\cdot||_2$ indicates a 2-norm operator, and $c:\mathbb{R}^d\rightarrow\mathbb{Z}$ is a mapping function that transforms the angle of inputs into integers based on hash coding  \cite{hashcode}. For hash coding, SimHash \cite{hash} is used to measure the approximate angle of the observation difference, and detail of hash coding is provided in Appendix A. Then, $D^{ij}$ contains the distance and angle information of $o^i-o^j$ in a much lower dimension.

Now, using the compressed difference $D^{ij}$, we define the formation $\mathcal{F}_{F^0,\cdots,F^{n-1}}$ of the exploration state $s^e = (o^0,~\cdots,~o^{n-1}) \in \mathcal{S}^e$ as follows:
\begin{equation}
\mathcal{F}_{F^0,\cdots,F^{n-1}}(s_e) = (\mathcal{F}_{F^0}^0, \cdots, \mathcal{F}_{F^{n-1}}^{n-1}),
\label{eq:formationtotal}
\end{equation}
where $n$ is the number of agents, $F^{i} = \{j_0,\cdots,j_{k-1}\} \subset I$ is an arbitrary ordered agent index set, and $\mathcal{F}^i_{F^i}$ is the individual formation viewed by $i$-th agent, defined as
\begin{equation}
\mathcal{F}^i_{F^i} = (D^{ij_0},\cdots,D^{ij_{k-1}}).
\label{eq:formationind}
\end{equation}
From the definition, the formation of the exploration state $s^e$ contains all the difference information of $(i,j)$ agent pairs for all $i\in I,~j\in F^i$. From now on, we denote the formation as $\mathcal{F}$ for better clarity.

Finally, we can define a formation-based binary relation on the exploration state space as
\begin{equation}
\sim_\mathcal{F}~:=\{(s_1,s_2)\in \mathcal{S}^e\times \mathcal{S}^e~|~ \mathcal{F}(s_1) = \mathcal{F}(s_2)\}, 
\end{equation}
and we can easily prove that $\sim_\mathcal{F}$ is an equivalence relation on the exploration space $\mathcal{S}^e$ by Lemma \ref{lem:equivalence}.
\begin{lemma}[Formation-based equivalence relation]
The binary relation $\sim_\mathcal{F}$ is an equivalence relation on the exploration space $\mathcal{S}^e$, i.e., two exploration states $s_1$ and $s_2$ in the exploration state $\mathcal{S}^e$ are equivalent under $\sim_\mathcal{F}$ if $\mathcal{F}(s_1) = \mathcal{F}(s_2)$.
\label{lem:equivalence}
\end{lemma}

Proof) Proof of Lemma \ref{lem:equivalence} is provided in Appendix B.

Here, the dimension of formation $\mathcal{F}$ is $nk$ and the dimension of exploration space is $nd$. $k$ depends on the selection of $F^{i}$, but in most cases, the observation dimension $d$ is much larger than $k$. Thus, we can efficiently reduce the dimension of search space. In addition, based on the equivalence relation $\sim_\mathcal{F}$, the agents explore the exploration states in diverse formations without visiting the equivalent states that have the same formation, and it can further reduce the search space and enables better exploration as shown in Figure \ref{fig:heatmap}.

\begin{figure}[t]
\centering
\includegraphics[width=\columnwidth]{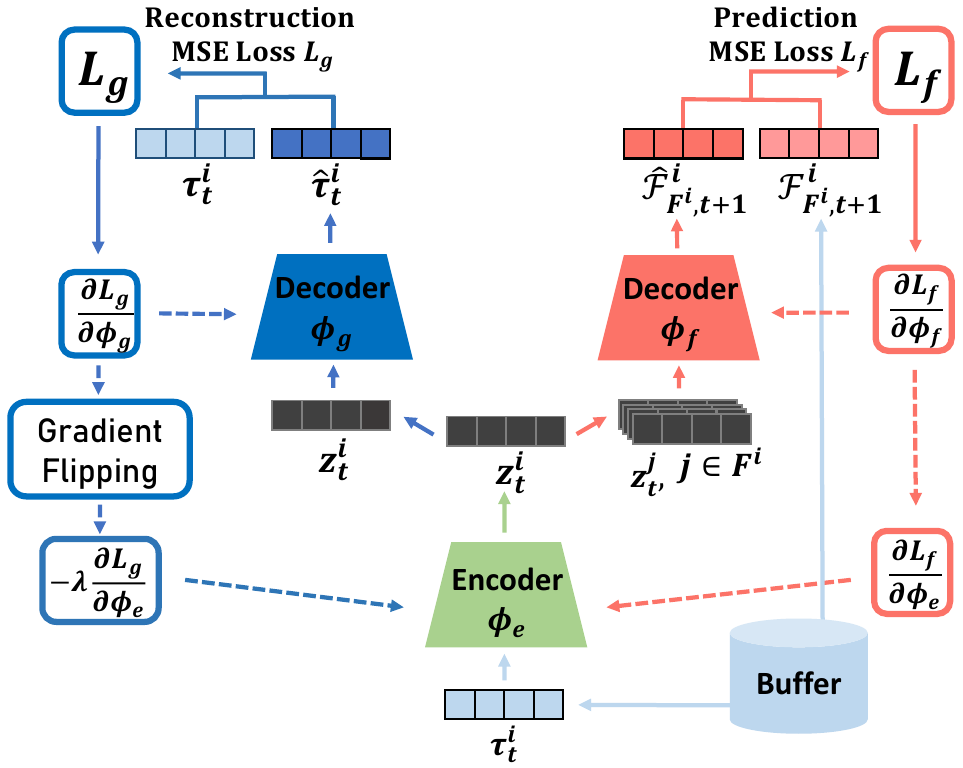} 
\caption{$\mathcal{F}$-Net Architecture}
\vspace{-.5em}
\label{fig:awarenet}
\end{figure}

\subsection{Formation-aware Exploration Objective}
In this subsection, we propose a formation-aware exploration objective to explore states in diverse formations as
{\small
\begin{equation}\nonumber
\underbrace{\mathcal{H}(\mathcal{F}_t)}_{(a)} + \frac{1}{n(k+1)}\sum_{i=0}^{n-1}\sum_{j\in F^{i+}} \underbrace{\mathcal{I}(\mathcal{F}^i_{F^i,t+1};z^j_t|\mathcal{\tau}^{F^{i+}}_t,z^{F^{i+}\backslash\{j\}}_t),}_{(b)}
\end{equation}}
where $F^{i+} = F^{i}\cup \{i\}$, $a^{F}$ indicates the set $\{a^i|i\in F\}$, $\mathcal{H}(X) = -p(X) \log p(X)$ is the information entropy of a random variable $X$, and $\mathcal{I}(X;Y) = \mathcal{H}(X) - \mathcal{H}(X|Y)$ is the mutual information between random variables $X$ and $Y$.

In the exploration objective, (a) is the entropy of formations, so maximizing (a) encourages the agents to visit states in diverse formations as illustrated in Figure \ref{fig:concepta}(a). Here, maximizing (a) can be accomplished by visitation count-based exploration \cite{vc2}. We define a formation-based count exploration reward $r^{exp}_t = \frac{1}{\sqrt{N(\mathcal{F}_t)}}$, then agents get higher rewards if they visit rarely visited formations. In order to count the number of formation $N(\mathcal{F}_t)$, we use $round(\cdot)$ method to discretize the formation $\mathcal{F}_t$, and then count the visitation number in the discretized bins. A detailed explanation for counting is provided in Appendix A. 
However, agents are partially observable in MARL, so visiting diverse formations by maximizing (a) can be a challenging problem if the agents do not know the formation information at all. 

For instance, the upper illustration of Figure \ref{fig:concepta}(b) depicts a situation where an agent lacks knowledge of the formation information. In the SMAC \cite{smac} environment, each agent is assigned a sight range. While an agent can observe other agents within its sight range, it remains uninformed about agents outside its sight range. Consequently, if the other agents are positioned outside its sight range, the agent may not be fully aware of the formation, as the formation includes information about all agents involved. Therefore, we consider an additional formation-aware objective (b), which is the mutual information between the latent variable $z_t^j,~\forall j \in F^{i+}$ and the next formation $\mathcal{F}_{F^i,t+1}^i$, where the trajectory $\tau_t^j$, the trajectories $\tau_t^{F^{i+}}\backslash\{j\}$ and the latent variables $z_t^{F^{i+}\backslash\{j\}}$ of other agents are given. Then, maximizing (b) guides the $i$-th agent to be aware of not only its own next formation $\mathcal{F}^i_{F^{i},t+1}$ but also all formations  $\mathcal{F}^j_{F^{j},t+1},$ $\forall j $ s.t. $i\in F^j$ to which $i$-th agent belongs as illustrated in Figure \ref{fig:concepta}(b). Thus, we maximize both (a) and (b) to make agents visit exploration states in diverse formations while recognizing their formations well.

In order to maximize (b), we derive an evidence lower bound for the mutual information term based on the variational inference approaches \cite{vae, cds}, and then we will maximize the evidence lower bound to increase the mutual information.
\begin{lemma} [Evidence lower bound]
Mutual information (b) can be lower-bounded as
\begin{equation}
\begin{aligned}
&\mathcal{I}(\mathcal{F}^i_{F^i,t+1};z^j_t|\mathcal{\tau}^{F^{i+}}_t,z^{F^{i+}\backslash\{j\}}_t) \geq \\ 
& \mathbb{E}_{\mathcal{F}^i_{F^i,t+1}, z^j_t\sim q_{\phi_e}} 
 \big[ \log q_{\phi_f}(\mathcal{F}^i_{F^i,t+1}|\mathcal{\tau}^{F^{i+}}_t,z^{F^{i+}}_t) \\
& \quad\quad\quad- \log p(\mathcal{F}^i_{F^i,t+1}|\mathcal{\tau}^{F^{i+}}_t,z^{F^{i+}\backslash\{j\}}_t)\big],
\end{aligned}
\label{eq:evidence}
\end{equation}
\label{lem:lemlb}
\end{lemma}
where $q_{\phi_f}(\cdot|\tau_t^{F^{i+}},z_t^{F^{i+}})$ is an arbitrary posterior distribution, and $\phi_f$ and $\phi_e$ parameterize the distributions $q_{\phi_e}$ and $q_{\phi_f}$, respectively.

Proof) Proof of Lemma \ref{lem:lemlb} is provided in Appendix B.

\begin{figure}[!t]%
    \centering
    \includegraphics[width=0.48\columnwidth]{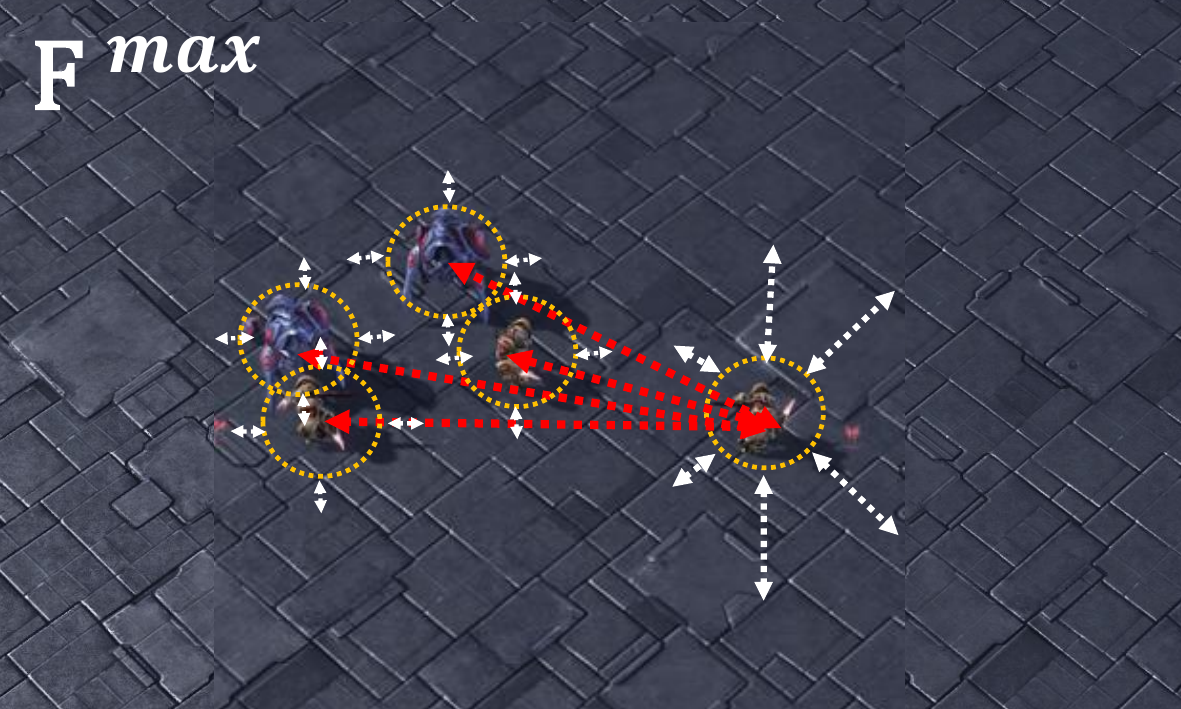}
    \includegraphics[width=0.48\columnwidth]{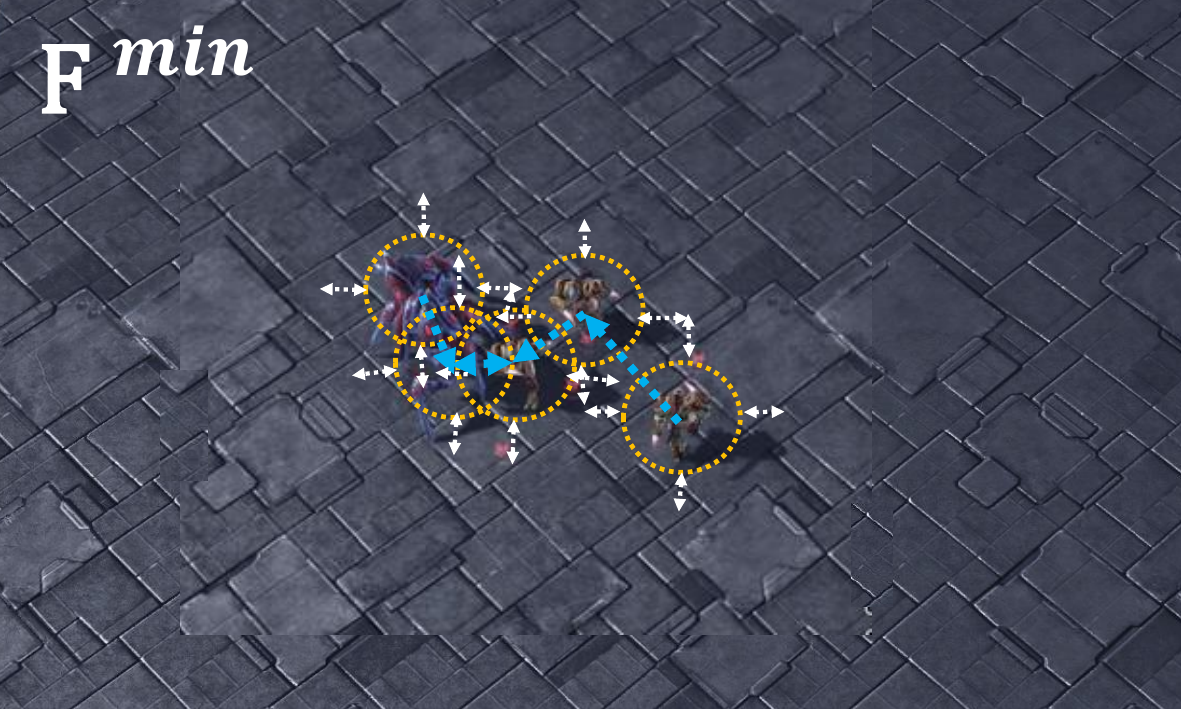} \\
    \vspace{1.5pt}
    \includegraphics[width=0.48\columnwidth]{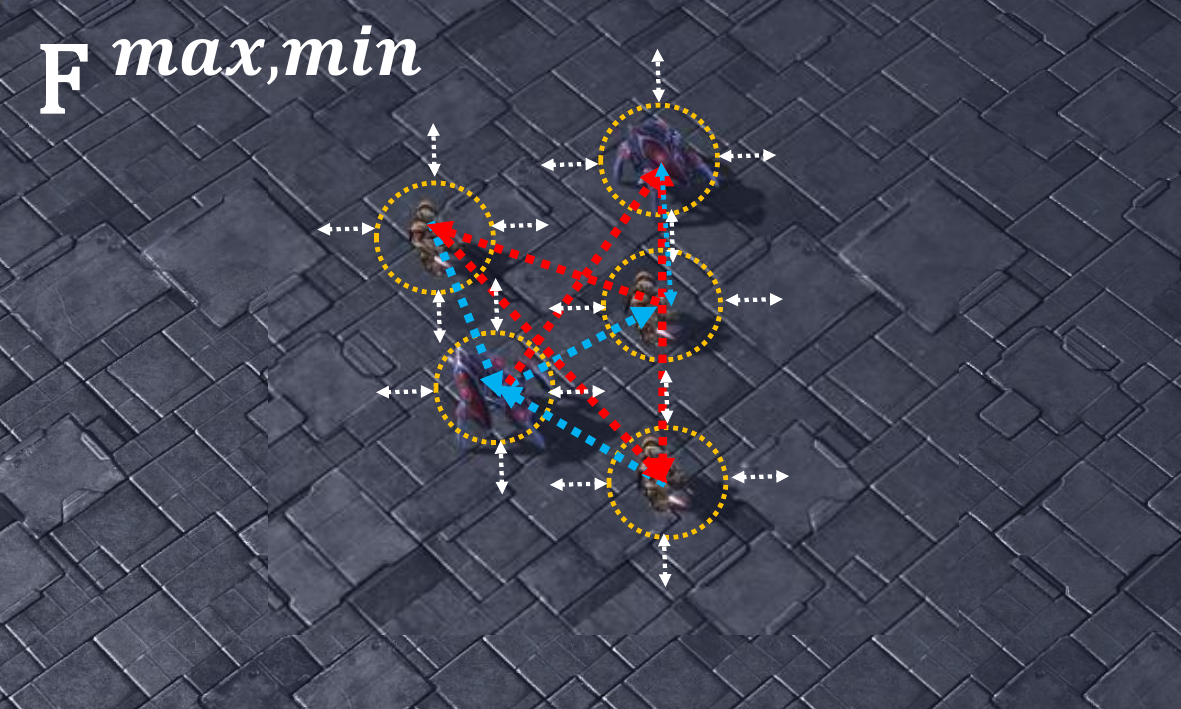}
    \includegraphics[width=0.48\columnwidth]{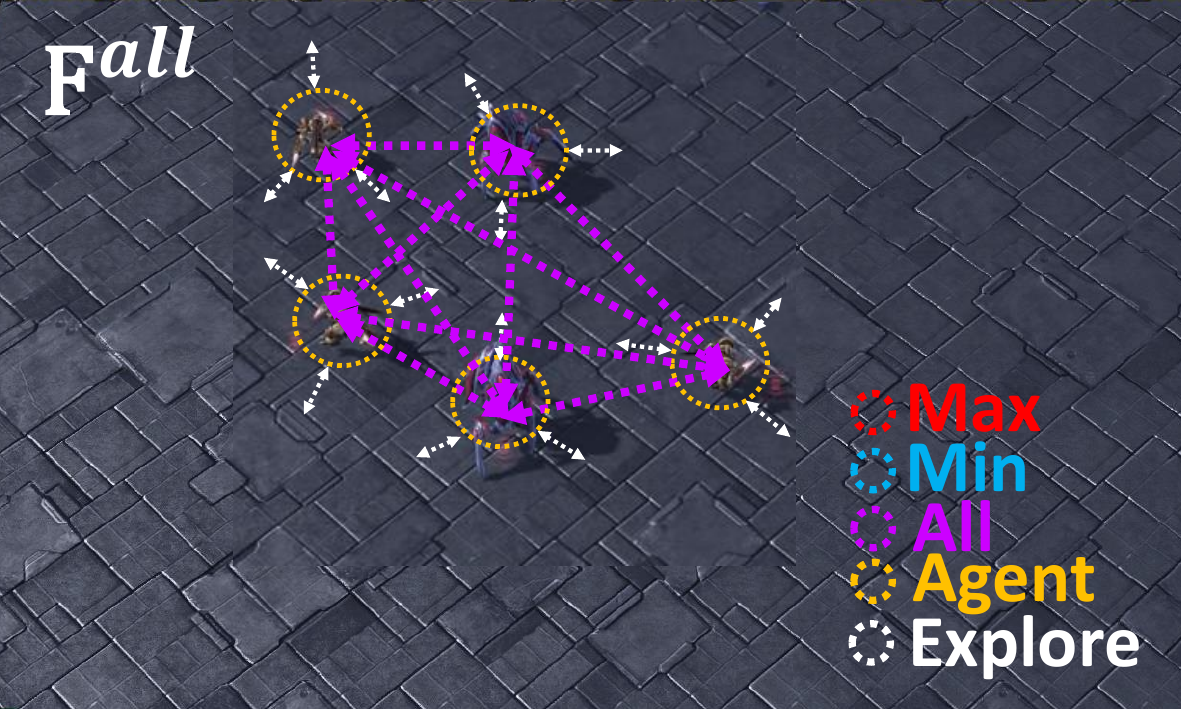}
    \caption{Formations with various agent index set $F$. While $F^{i, max}$, $F^{i, min}$ focuses on particular agent relationships, $F^{i,all}$ and $F^{i,max,min}$ considers extensive agent relationships.}
    \label{fig:formation}
    \vspace{-.5em}
\end{figure}

In the proof of Lemma \ref{lem:lemlb}, $z_t^j$ is averaged over the distribution $q_{\phi_e}(\cdot|\tau_t^{F^{i+}}, z_t^{F^{i+}\backslash\{j\}})$, but it makes difficult to utilize the latent variable $z_t^j$ in the decentralized execution setup since $j$-th agent cannot exploit the other agent's trajectory information. Thus, for practical implementation, we consider the approximate distribution $q_{\phi_e}(\cdot|\tau_t^j)$ instead of $q_{\phi_e}(\cdot|\tau_t^{F^{i+}}, z_t^{F^{i+}\backslash\{j\}})$. Then, we can formulate the encoder-decoder structure such as variational auto-encoder (VAE) \cite{vae}, and we define an encoder $E_{\phi_e}(\tau_t^i)$ whose outputs are the mean and standard deviation of the latent variable $z_t^i$ to sample the latent variable and a decoder $D_{\phi_f}(z_t^{F^i+})$ whose output is the prediction of next formation $\hat{\mathcal{F}}^{i}_{F^i,t+1}$.
Here, note that we exploit the agent's trajectory $\tau_t^i$ to extract $z_t^i$ from the encoder, but the trajectory $\tau_t^i$ may contain the redundant information regardless of the formation prediction, and it can be delivered to the latent variable $z_t^i$. The irrelevant information can disturb formation-aware exploration as in the case of \citet{curi2}. Thus, we aim to consider a gradient flipping (GF) technique proposed in the field of domain adaptation \cite{unsupervised} to remove the redundant information. In order to apply GF in our setup, we consider an additional trajectory decoder $D_{\phi_g}(z^i_t)$ whose output is the reconstruction of trajectory $\hat{\tau}^i_t$ and we update the encoder parameter $\phi_e$ to prevent the trajectory reconstruction, while decoder strives to reconstruct the trajectory. Then, we can prevent delivering useless information from the trajectory $\tau_t^i$ to the latent variable $z_t^i$ as proposed in \citet{unsupervised}.

In summary, we can construct $\mathcal{F}$-Net architecture as shown in Figure \ref{fig:awarenet}. We define a formation prediction loss $L_f(\phi_e,\phi_f) = \frac{1}{n}\sum_{i=0}^{n-1} MSE (\mathcal{F}^{i}_{F^i, t+1},~ \hat{\mathcal{F}}^i_{F^i,t+1})$, a trajectory reconstruction loss $L_g(\phi_e,\phi_g) = \frac{1}{n}\sum_{i=0}^{n-1} MSE (\tau^{i}_t,~ \hat{\tau}^i_{t})$, and Kullback-Leibler divergence loss $L_{KL}(\phi_e)=D_{KL}(q_{\phi_e}(\cdot|\tau_t^i)||\mathcal{N}(0,\mathbf{I}))$ as in \citet{vae,unsupervised}, where $MSE(x,y)=\mathbb{E}[(x-y)^2]$ is the mean square error loss and $\mathcal{N}(0,\mathbf{I})$ is the multi-variate standard normal distribution with identity matrix $\mathbf{I}$. Based on the loss functions, we can update the encoder-decoder parameters $\phi_e,~\phi_f,~\phi_g$ as follows:
\begin{equation}
\begin{aligned}
\phi_e &\leftarrow (1-\alpha) \phi_e + \alpha \left(\frac{\partial L_f}{\partial\phi_e} + \frac{\partial L_{KL}}{\partial\phi_e}-\lambda_{GF}\frac{\partial L_g}{\partial\phi_e}\right)\\
\phi_f &\leftarrow (1-\alpha) \phi_f + \alpha \frac{\partial L_f}{\partial\phi_f}, ~\phi_g \leftarrow (1-\alpha) \phi_g + \alpha \frac{\partial L_g}{\partial\phi_g},
\end{aligned}
\label{eq:netupdate}
\end{equation}
where $\alpha$ is a learning rate, $\lambda_{GF}$ is a hyperparameter that controls the gradient flipping effect, and we fix $\lambda_{GF}=0.1$ in our setup. Note that $\phi_e$ maximizes the reconstruction loss $L_g$ to prevent the trajectory reconstruction. 

Finally, in order to maximize the evidence lower bound in \eqref{eq:evidence} to be aware of the formation information, we design the intrinsic reward $r^{aware}$ that approximately represents the evidence lower bound as follows:
\begin{equation}
\begin{aligned}
    & r_t^{aware} = \frac{1}{n}\sum_{i=0}^{n-1}\bigg(\log q_{\phi_f} (\hat{\mathcal{F}}^i_{t+1}|\tau^{F^{i+}}_t,z^{F^{i+}}_t) \\
    & - \frac{1}{k+1}\sum_{j \in F^{i+}} \log p
    (\hat{\mathcal{F}}^i_{t+1}|\tau^{F^{i+}}_t,z^{F^{i+}\backslash\{j\}}_t)\bigg),
\end{aligned}
\label{eq:rewaware}
\end{equation}
the detailed implementation of $r_t^{aware}$ using formation prediction loss is provided in Appendix C. 

\begin{figure}[t]
\centering
\includegraphics[width=\columnwidth]{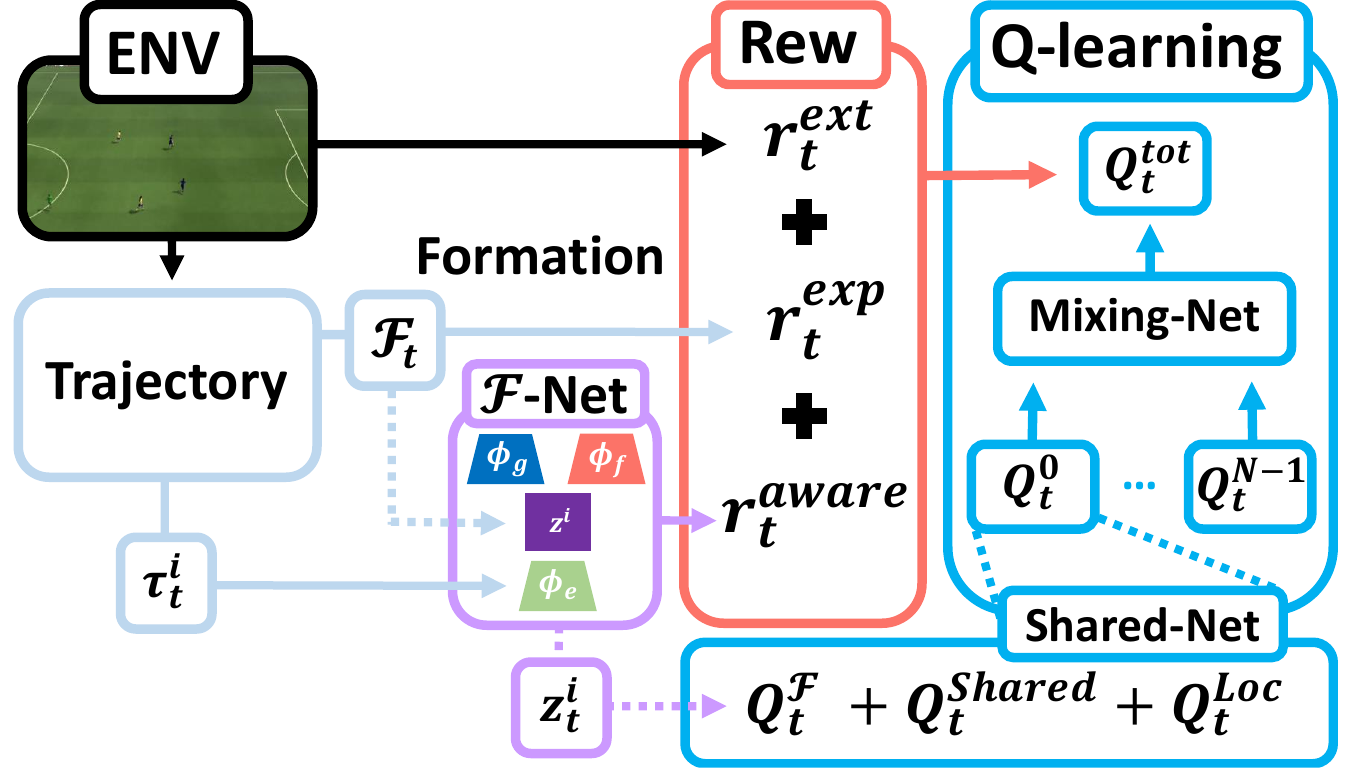}
\caption{Schema of FoX framework}
\label{fig:schema}
\vspace{-.5em}
\end{figure}

Finally, with the extrinsic reward $r_t^{ext}$ given from the environment and intrinsic rewards $r_t^{exp}$ and $r_t^{aware}$ for formation-aware exploration, we can define the total reward $r^{tot}_t = r_t^{ext} + \beta_1 r_t^{exp} + \beta_2 r_t^{aware}$, where $\beta_1$ and $\beta_2$ are hyperparamers to control the effect of our intrinsic rewards $r_t^{exp}$ and $r_t^{aware}$, respectively. Then, an overall temporal difference loss to update the total $Q$-function $Q_\theta^{tot}$ in \eqref{eq:qmix} parameterized by $\theta$ is defined as 
\begin{equation}
\begin{aligned}
L_{TD}(\theta) &= (r_t^{tot} + \gamma \max_{a'} Q_{\theta^-}^{tot}(s_{t+1},a') - Q^{tot}_{\theta}(s_t,a_t))^2
\end{aligned}
\label{eq:tdloss}
\end{equation}
where $\theta^-$ is a parameter for the target network updated by the exponential moving average (EMA). We summarize the proposed FoX framework as Algorithm \ref{alg:algorithm} and Figure \ref{fig:schema}.

\subsection{Selection of Index Set $F^i$}
In arranging formations, various agent index sets $F^i$ can be considered. For example, $F^{i,max}:= \{ \argmax_{j \in I \backslash \{i \}} d(i,j)\}$ forms a formation among agents with the biggest difference. In our experiments, we observe the exploration behaviors of various formations based on $F^{i,max}$, $F^{i,min} = \{ \argmin_{j \in I \backslash \{i \}} d(i,j)\}$, $F^{i,max,min} = F^{i,max} \cup F^{i,min}$, and $F^{all} = \{i|i \in I \backslash \{i \} \}$ which is illustrated in Figure \ref{fig:formation}. We conduct an ablation study for the selection of $F^i$, and the result shows that $F^{i,max,min}$ shows the best performance.

\subsection{Formation-based Shared Network}
Based on the shared network structure for individual $Q$-function proposed in \citet{cds}, we aim to exploit the latent variable $z_t^i$. Thus, individual $Q$-function $Q^i$ in \eqref{eq:qmix} can be decomposed as three $Q$ functions as follows:
\begin{equation}
    Q^i(\tau_t^i,z_t^i,\cdot)= Q^{Shared}(\tau_t^i,\cdot) + Q^{Loc,i}(\tau_t^i,\cdot) + Q^{\mathcal{F}}(z_t^i,\cdot),
\end{equation}
where $Q^{Shared}$ is a shared Q-function, $Q^{Loc,i}$ is $i$-th local $Q$-function, and $Q^{\mathcal{F}}$ is a $Q$-function that exploits formation information using $z_t^i$. As proposed in \citet{cds}, we also consider $l1$ regularization for $Q^{Loc,i}$ with the regularization coefficient $\lambda_{reg}=0.1$.

\begin{algorithm}[t]
\caption{FoX framework}
\label{alg:algorithm}
Initialize $\phi_e$, $\phi_f$, $\phi_g$, $\theta$, $\theta^-$
\begin{algorithmic}[1] 
\STATE \textbf{for} each epoch \textbf{do}\\
\STATE \ \ \ \ \textbf{for} each gradient step \textbf{do}
\STATE \ \ \ \ \ \ \ \ Obtain trajectory samples from environment
\STATE \ \ \ \ \ \ \ \ Choose random batch from buffer $\mathcal{D}$
\STATE \ \ \ \ \ \ \ \ Calculate formation $\mathcal{F}_t$ from $\textbf{o}_t$ in \eqref{eq:formationtotal}
\STATE  \ \ \ \ \ \ \ \ Compute intrinsic rewards $r^{exp}$, $r^{aware}$
\STATE  \ \ \ \ \ \ \ \ Compute loss functions $L_f$, $L_g$, $L_{KL}$, $L_{TD}$
\STATE  \ \ \ \ \ \ \ \ Update $\mathcal{F}$-Net 
 parameters $\phi_e$, $\phi_f$, $\phi_g$ based on \eqref{eq:netupdate}
\STATE  \ \ \ \ \ \ \ \ Update $Q$-function parameter $\theta$
\STATE \ \ \ \ \ \ \ \ Update target parameter $\theta^{-}$ using EMA.
\STATE \ \ \ \ \ \ \ \ Store samples in buffer $\mathcal{D}$
\STATE \ \ \ \ \textbf{end for}
\STATE \textbf{end for}
\end{algorithmic}
\end{algorithm}

\begin{figure*}[!ht]%
    \centering
    \includegraphics[width=0.3\textwidth]{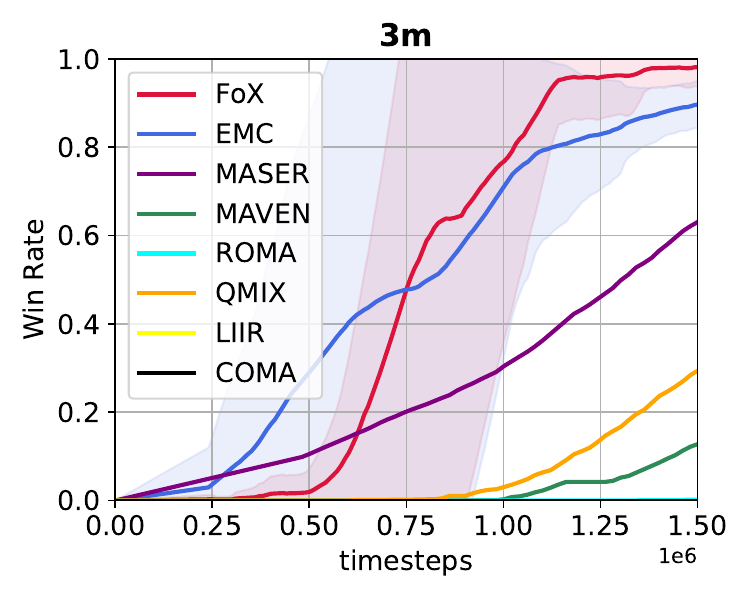}
    \includegraphics[width=0.3\textwidth]{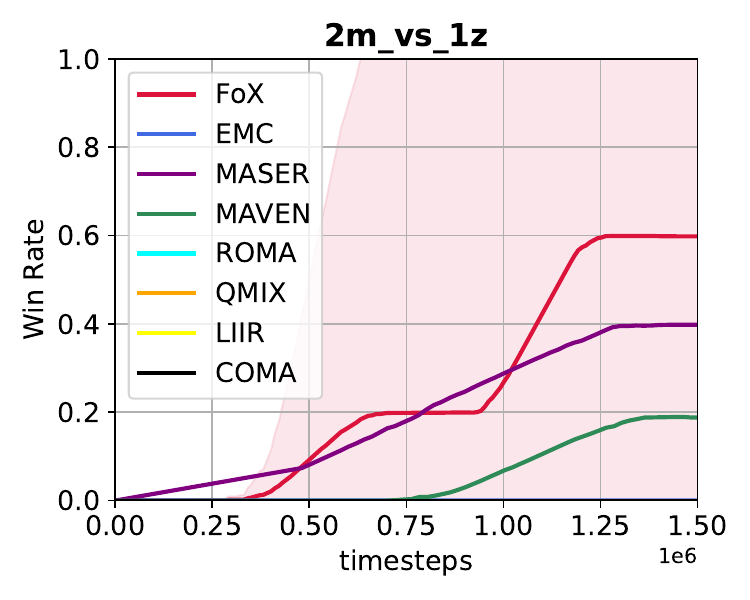}
    \includegraphics[width=0.3\textwidth]{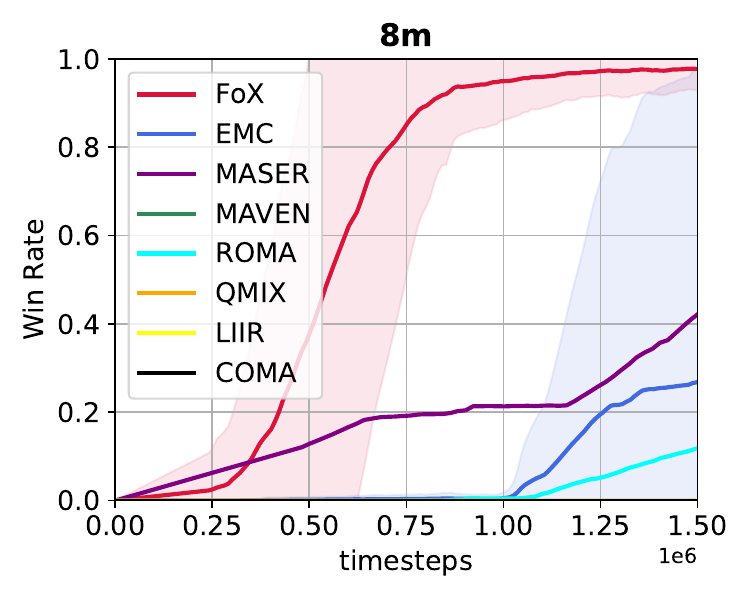}
    \includegraphics[width=0.3\textwidth]{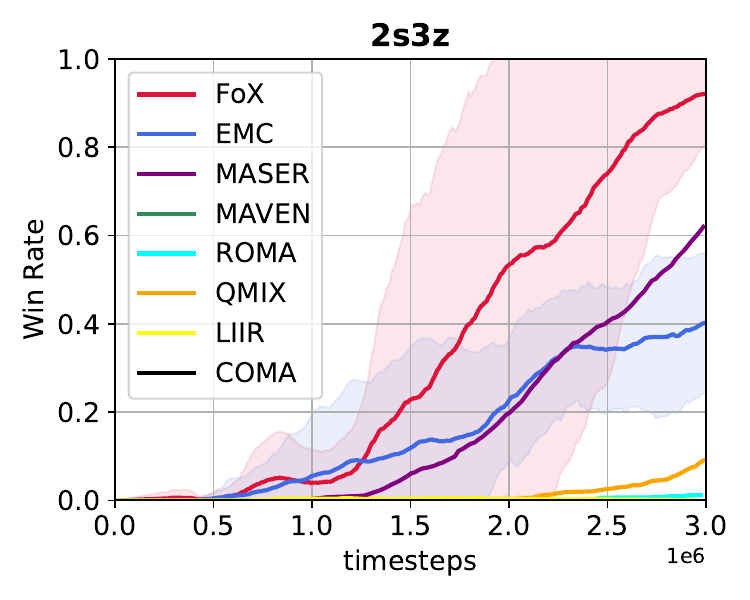}
    \includegraphics[width=0.3\textwidth]{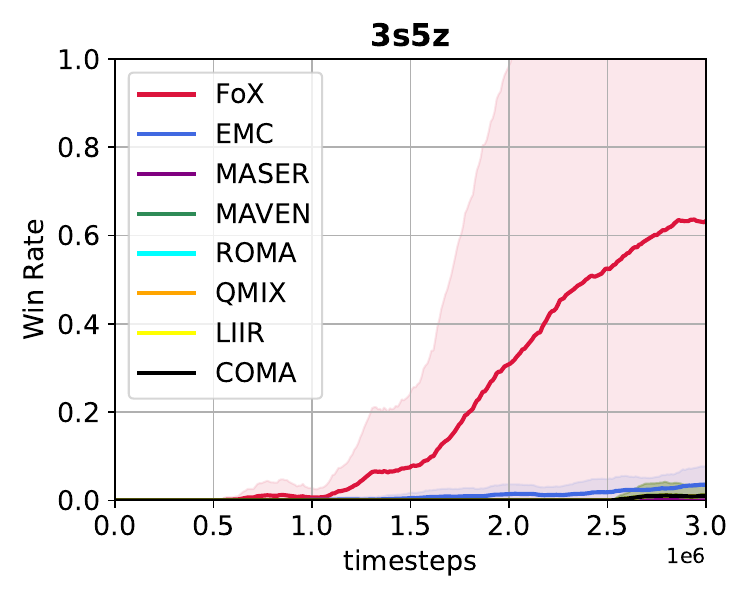}
    \includegraphics[width=0.3\textwidth]{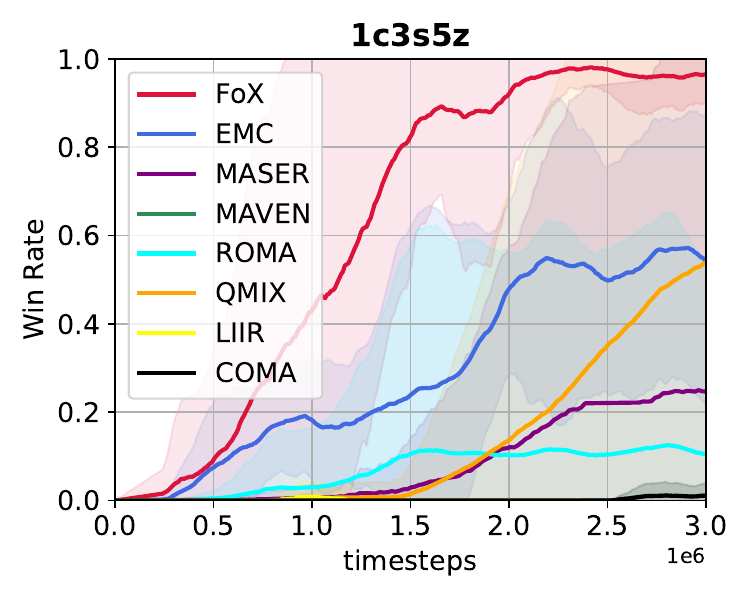}
    \caption{Performance results on SMAC(sparse)}
    \label{fig:smac}
\end{figure*}

\begin{figure*}[!ht]%
    \centering
    \includegraphics[width=0.24\textwidth]{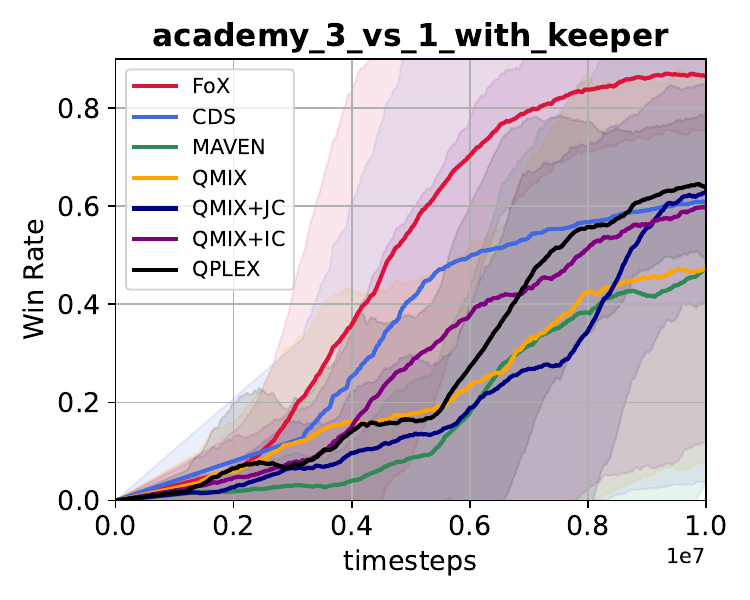}
    \includegraphics[width=0.24\textwidth]{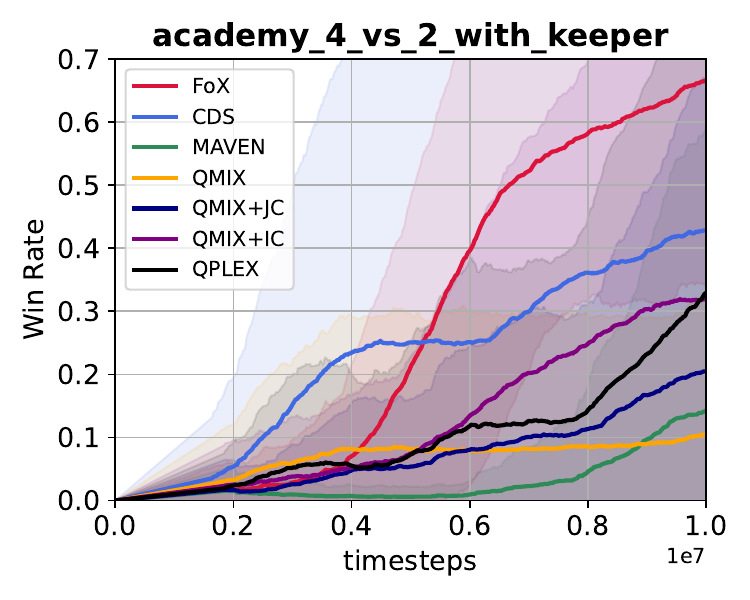}
    \includegraphics[width=0.24\textwidth]{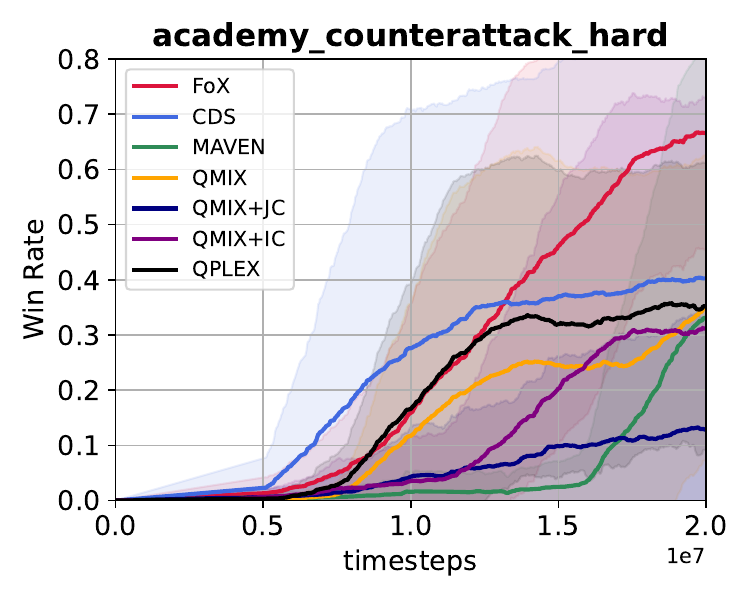}
    \includegraphics[width=0.24\textwidth]{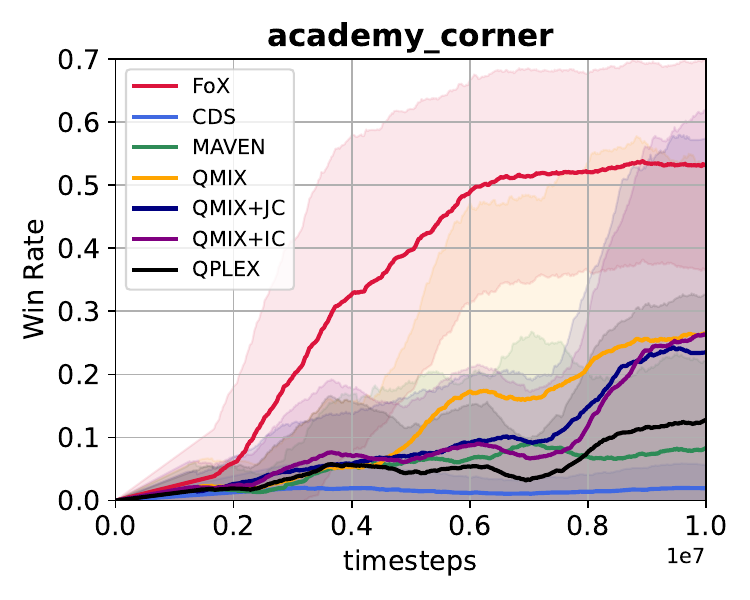}
    
    \caption{Performance results on GRF}
    \label{fig:grf}
\end{figure*}

\section{Experiments}
In our experiments, we evaluate the performance of the proposed FoX framework in the challenging cooperative multi-agent environments: the sparse StarCraft multi-agent challenge (SMAC) \cite{smac} and Google Research Football (GRF) \cite{grf}. The source code of our proposed algorithm is available at https://github.com/hyeon1996/FoX. Detailed settings of the environment can be found in D. In the SMAC and GRF environments, the observation of an agent or enemy force identified for elimination or out of sight is automatically set to zero or remains constant. Thus the difference between observation of such agent remain unchanged, becoming ineffective in formation computation. Therefore, in the SMAC and the GRF environment, our formation can effectively represent the topological changes. For hyperparameters $\beta_1$, $\beta_2$, we have tested $\beta_1 \in\{ 0.001, 0.005, 0.01, 0.02, 0.1\}$ and $\beta_2 \in \{ 0.001, 0.005, 0.01, 0.05\}$ for both environments. We used the best parameters for each environment to demonstrate our experimental results. Detailed hyperparameter setups can be found in Appendix E. We averaged 5 seeds for SMAC tasks and 5 seeds for GRF tasks, and the solid line in the performance graph indicates the average performance, and the shaded part represents the standard deviation. 

\subsection{Performance on Sparse SMAC}
In the original scenarios of SMAC, agents are densely rewarded for the damage they have dealt or taken in addition to sparse rewards upon the deaths of an ally or the enemy and defeating the enemy. In the sparse SMAC environment, the agents must learn with only sparse rewards without the dense rewards thus making the task more challenging. Detailed reward settings of the dense and sparse SMAC can be found in Table \ref{tab:smacr}. To leverage the fact that formation-aware exploration is not value-dependent, we conducted experiments in sparse environments. In sparse SMAC, we evaluate the performance of FoX in 6 scenarios: 3m, 8m, 2s3z, 2m\_vs\_1z, 3s5z, and 1c3s5z. While the 2m\_vs\_1z scenario has the fewest number of agents with $n=2$, the 1c3s5z scenario has the highest number of agents with $n=9$. 

\begin{table}[ht]
  \centering
  \small
  \begin{tabular}{ccc}
    \toprule
    \textbf{Event}& \textbf{Dense reward} & \textbf{Sparse reward} \\
    \midrule
    Death of an enemy & +10 & +10 \\
    Death of an ally & -5 & -5 \\
    Win & +200 & +200 \\
    Enemy hit-point & -Enemy hit-point & - \\
    Ally hit-point & +Ally hit-point & - \\
    Other elements & +/- other components & - \\
    \bottomrule
  \end{tabular}
  \caption{The reward setting for dense and sparse SMAC.}
  \label{tab:smacr}
\end{table}

The exploration scheme of FoX can be easily applied to other existing MARL algorithms, but as we implement Fox on QMIX \cite{QMIX} during our experiments, we tested the performance of FoX against several QMIX-based algorithms including EMC \cite{emc}, MASER \cite{maser}, and ROMA \cite{roma}. EMC employs exploration based on the prediction error of agent influence, MASER assigns subgoals with the highest individual and joint Q value, and ROMA assigns roles based on agent trajectories. FoX was also compared with MAVEN \cite{maven} in the context of exploration via variational inference, LIIR \cite{liir} for intrinsic reward learning, and COMA \cite{coma} as a policy-gradient method. All experiments were conducted with the author-provided codes. As both FoX and EMC can be implemented on QMIX or QPLEX \cite{qplex}, we conduct our experiments with QMIX-based FoX and EMC for fair comparison.
\begin{figure}[ht]
\centering
\includegraphics[width=0.495\columnwidth]{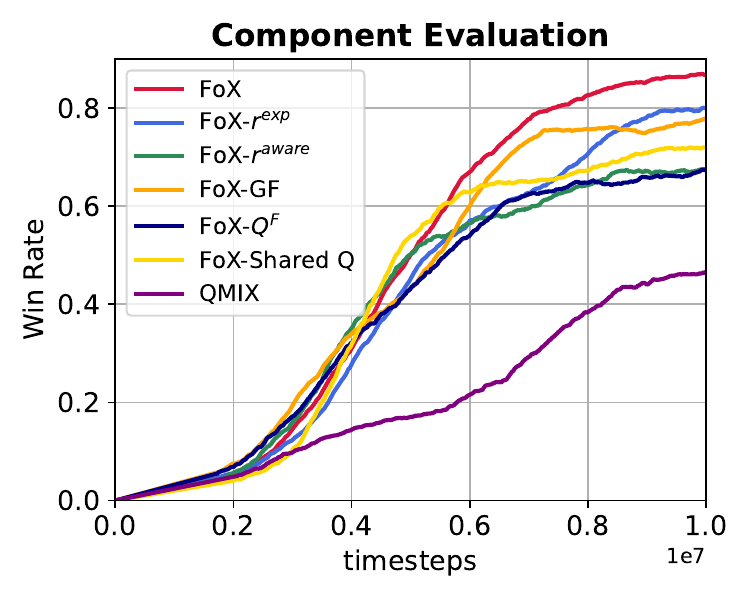}
\includegraphics[width=0.495\columnwidth]{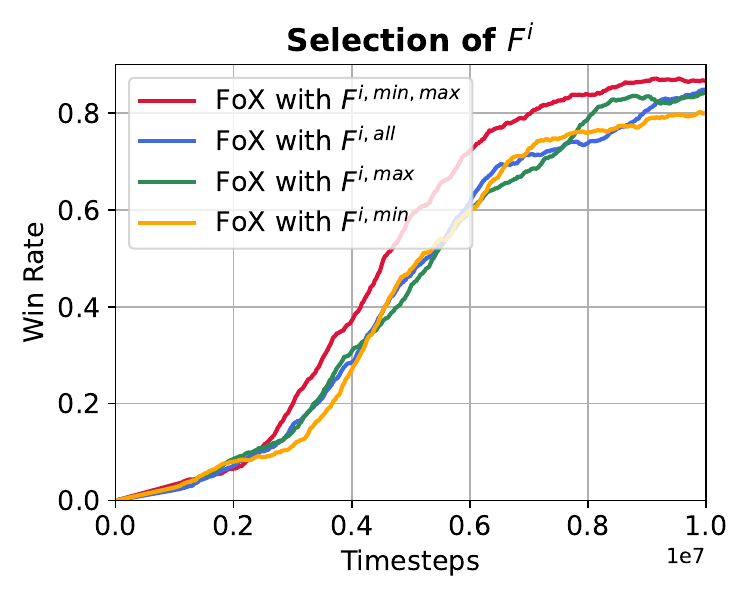}
\includegraphics[width=0.495\columnwidth]{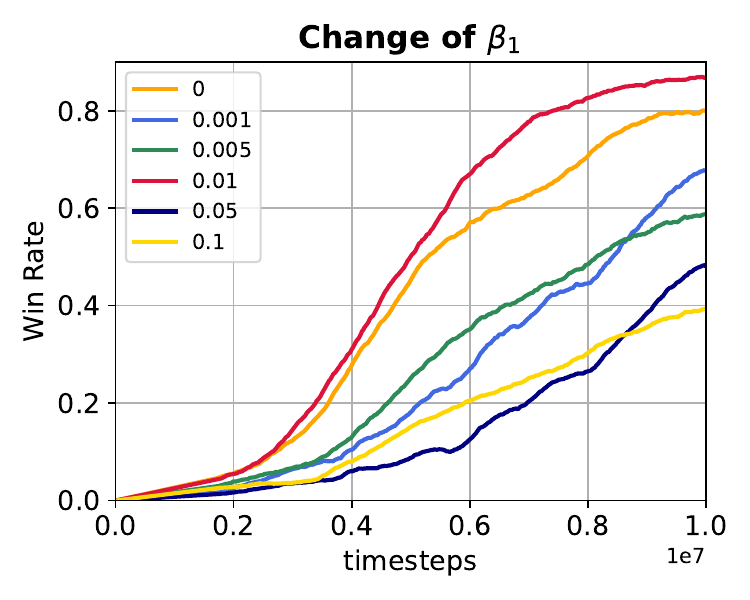}
\includegraphics[width=0.495\columnwidth]{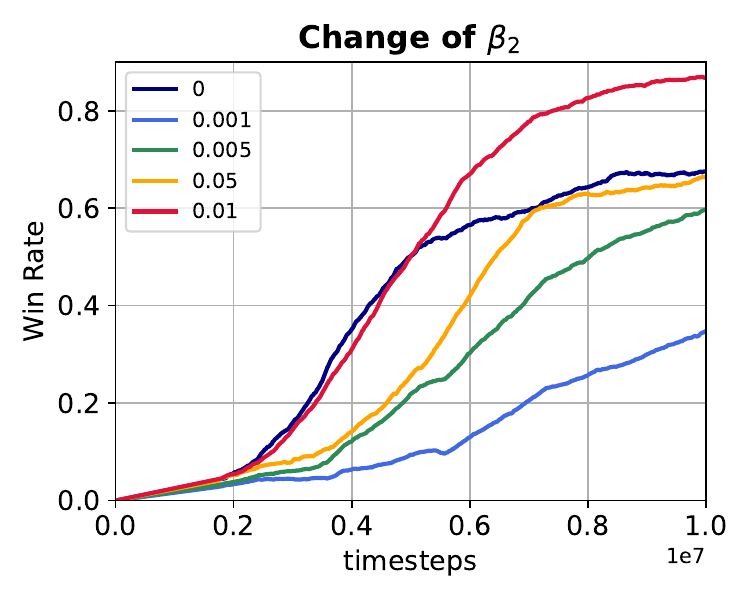}
\caption{Ablation studies on GRF}
\label{fig:comp}
\end{figure}
From the experimental results, we can see that in the relatively easier environment of 3m, EMC shows comparable results to FoX. However, FoX significantly outperforms the other baselines in the other scenarios, where the results in 2m$\_$vs$\_$1z are especially notable as the other baselines struggled to learn the environment.

\subsection{Performance on GRF}
In Google Research Football \cite{grf} the agents' observation contains the location of the players, the opponents, and the ball. In our implementation, the episode is terminated and a negative reward is given to the agents when the ball goes to the other half of the court. The reward for the GRF environment is described in Table \ref{tab:grfr}. In GRF, we compare the FoX framework with baselines such as QPLEX \cite{qplex}, CDS \cite{cds}, MAVEN \cite{maven}, and QMIX \cite{QMIX}. QPLEX incorporates the IGM principle into a neural network architecture, utilizing a duplex dueling structure, CDS allows the agents to adaptively decide whether to pursue diversity or share information based on agent trajectories. In addition, we also compare the visitation count method implemented on QMIX to show the efficiency of our formation-based exploration. The variations of QMIX are denoted as QMIX+JC for QMIX with joint observation visitation counts and QMIX+IC for the individual observation visitation counts. In our experiments, we evaluate the performance of FoX on four academy scenarios: 
$Academy\_3\_vs\_1\_with\_keeper$, $Academy \_4\_vs\_2\_with\_keeper$, $Academy\_counteratta$ $ck\_hard$, and $Academy\_corner$. While the $Academy\_$ $3\_vs\_1\_with\_keeper$ scenario has the least number of agents with $n=3$, the $Academy\_corner$ scenario has the most number of agents with $n=10$, corresponding to all ally players except the goalie. Figure 7 illustrates the performance of FoX and the baseline algorithms on the GRF scenarios. FoX shows outstanding performance in all four scenarios of the experiment, outperforming the other baselines. Especially, FoX outperforms the other visitation count baselines, supporting the idea that formation-based state equivalence efficiently reduces the search space in MARL. In addition, the exploration path of FoX agents in the GRF environment can be found in Appendix F, showing how agents become aware of their formation as the learning step increases.

\begin{table}[!ht]
  \centering
  \small
  \begin{tabular}{ccc}
    \toprule
    \textbf{Event}& \textbf{Checkpoint reward} & \textbf{Score reward} \\
    \midrule
    Our team scores & - & +1 \\
    Opposing team scores & - & -1 \\
    Score from checkpoint & +0.1 $\cdot$ checkpoints& - \\
    \bottomrule
  \end{tabular}
  \caption{The reward setting for GRF.}
  \label{tab:grfr}
\end{table}

\section{Ablation Study}
In this section, we analyze the contributions of FoX. As $\beta_1 = 0.01$ and $\beta_2 = 0.01$ showed the best performance among all GRF scenarios, we conduct our study in the GRF $Academy\_3\_vs\_1\_with\_keeper$ scenario with a default setting of $\beta_1,\beta_2 = 0.01$ in our ablation studies.
\subsection{Component Evaluation}
To measure the effects of each component in the FoX framework, we remove each component from the FoX algorithm. Component evaluation graph in Figure \ref{fig:comp} shows the effect of various components of the intrinsic rewards $r^{exp}$ and $r^{aware}$ for formation-aware exploration, the gradient flipping (GF), $Q$-function $Q^{\mathcal{F}}$ with formation information, and shared $Q$ learning. In Figure \ref{fig:comp}, FoX-$r^{exp}$ and FoX-$r^{aware}$ both show a decrease in their performance, so it shows that the intrinsic rewards are crucial components for improving the performance in FoX. Furthermore, it is noticeable that excluding $Q^{\mathcal{F}}$ and Shared $Q$ significantly reduces the performance.  Lastly, the importance of minimizing irrelevant information from individual trajectories upon latent variable extraction is shown through FoX-GF.  As gradient flipping allows $z$ to capture only the formation relevant information from $\tau$, removing gradient flipping shows a decrease in performance. 
\subsection{The Effect of Intrinsic Rewards}
As seen in the graph in Figure \ref{fig:comp}, formation-aware exploration proves its effectiveness in the $Academy$ $\_3\_vs\_1\_with\_keeper$ environment since $\beta_1=0.01$ works better than $\beta_1=0$, but excessive exploration with $\beta_1>0.01$ can slow down the learning for excessive exploration. Similarly, guiding the agents to be formation-aware helps the learning of the environment. According to the change in reward when $\beta_2 > 0.01$, the agents are guided to be aware of the formation. Awareness of the formation formulated by the other agents mitigates the information bottleneck from partial observability, boosting the learning process. 
\subsection{Formation Selection}
According to Figure \ref{fig:comp}, formation-based on $F^{i,max}$ and $F^{i,min}$ may hold enough information to represent the agent relationship since $Academy\_3\_vs\_1\_with\_keeper$ presents a relatively small number of agents $n=3$. Surprisingly, $F^{all}$ showed a decrease in performance even though it should hold a similar amount of information to formation based on $F^{i,max,min}$. Such behavior emphasizes the importance of formation selection. To see the effect of formation selection we have to consider the environment with a large number of agents such as 2s3z as represented in Figure \ref{fig:formation}.

\section{Conclusion}
Exploration space which can increase exponentially upon the number of agents makes efficient exploration in MARL very challenging. In this paper, we reduce the exploration space while capturing the condensed information of the state through formations, ensuring visitation to diverse formations with awareness. Finally, the efficiency of formation-aware exploration is demonstrated in the StarCraft II Multi-Agent Challenge and Google Research FootBall, where FoX shows state-of-the-art performance. 

\section*{Acknowledgement}
This work was supported in part by Institute of Information \& Communications Technology Planning \& Evaluation (IITP) grant funded by the Korea government (MSIT) (No.2022-0-00469, Development of Core Technologies for Task-oriented Reinforcement Learning for Commercialization of Autonomous Drones) and in part by IITP grant funded by the Korea government (MSIT) (No. RS-2022-00156361, Innovative Human Resource Development for Local Intellectualization(UNIST)) and in part by Artificial Intelligence Graduate School support (UNIST), IITP grant funded by the Korea government (MSIT) (No.2020-0-01336).

\bigskip

\clearpage

\bibliography{aaai24}

\end{document}